\pdfoutput=1

\documentclass[11pt]{article}

\usepackage{EMNLP2022}

\usepackage{times}
\usepackage{latexsym}

\usepackage[T1]{fontenc}

\usepackage[utf8]{inputenc}

\usepackage{microtype}

\usepackage{booktabs}

\usepackage{enumitem}

\usepackage{inconsolata}

\usepackage{graphicx}

\newcommand\sect[1]{\S\ref{#1}}

\title{
Calibrating Trust of Multi-Hop Question Answering Systems with Decompositional Probes
}

\author{Kaige Xie\textsuperscript{$\clubsuit$} \quad Sarah Wiegreffe\textsuperscript{$\dagger$} \quad Mark Riedl\textsuperscript{$\clubsuit$} \\
\textsuperscript{$\clubsuit$} School of Interactive Computing, Georgia Institute of Technology \\
\textsuperscript{$\dagger$}Allen Institute for Artificial Intelligence \\
\texttt{\{kaigexie, riedl\}@gatech.edu} \\
\texttt{wiegreffesarah@gmail.com}\\}

\begin{document}
\maketitle
\begin{abstract}
Multi-hop Question Answering (QA) is a challenging task since it requires an accurate aggregation of information from multiple context paragraphs and a thorough understanding of the underlying reasoning chains. Recent work in multi-hop QA has shown that performance can be boosted by first decomposing the questions into simpler, single-hop questions. In this paper, we explore one additional utility of the multi-hop decomposition from the perspective of explainable NLP: to create explanation by \textit{probing} a neural QA model with them. We hypothesize that in doing so, users will be better able to predict when the underlying QA system will give the correct answer. Through human participant studies, we verify that exposing the decomposition probes and answers to the probes to users can increase their ability to predict system performance on a question instance basis. We show that decomposition is an effective form of probing QA systems as well as a promising approach to explanation generation. In-depth analyses show the need for improvements in decomposition systems.~\footnote{Our code and data are available at \url{https://github.com/kaigexie/decompositional-probing}.}
\end{abstract}

\section{Introduction}
\label{sec:introduction}

\begin{figure}[t!]
\centering
  \includegraphics[width=\linewidth]{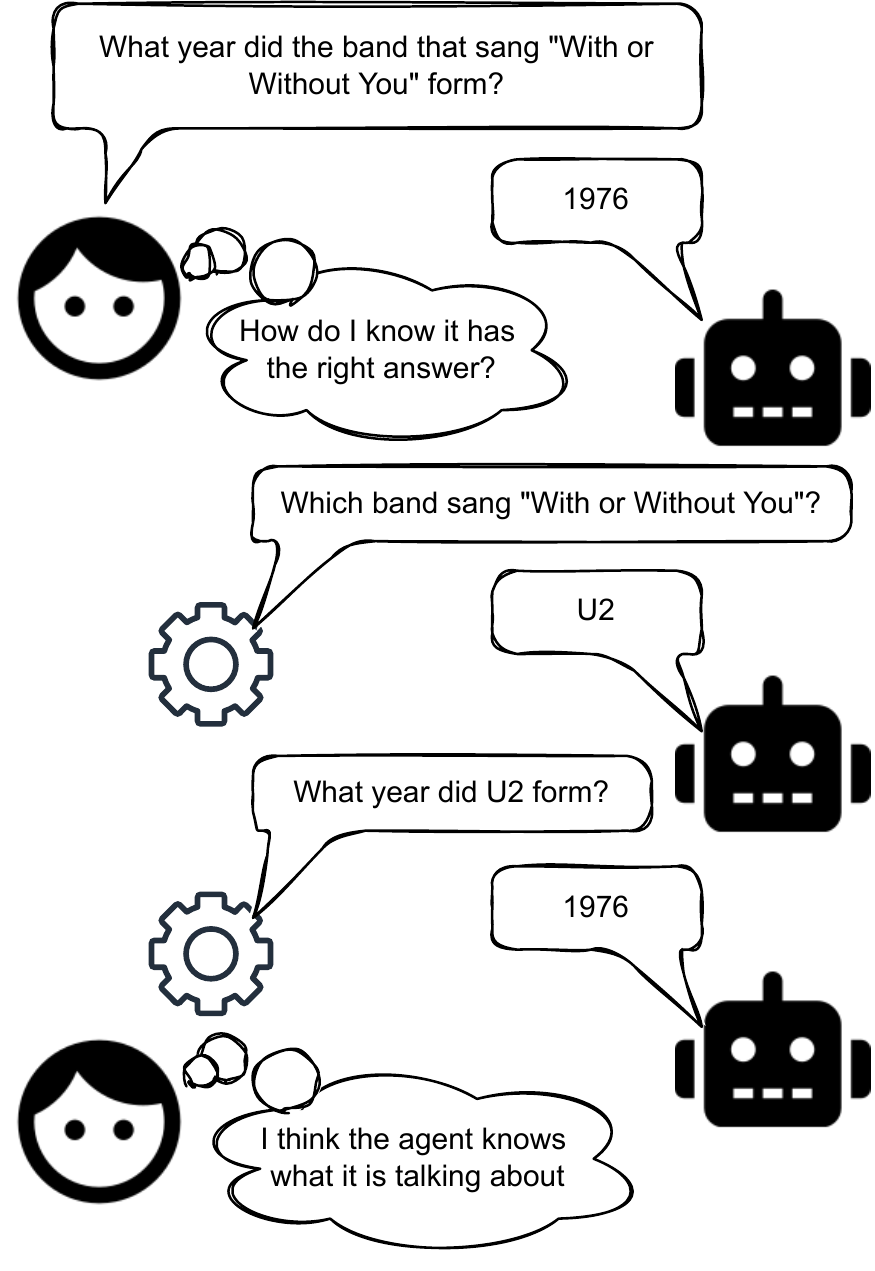}
\caption{An overview of our method. Users wonder if they are able to trust the answer. Sub-questions are generated by a decomposer agent (gear) to probe the question-answering agent.}
\label{fig:summary}
\end{figure}

As natural language understanding tasks have become increasingly complex, the field of explainable natural language processing (exNLP) aims to help users understand the performance of NLP systems. Multi-hop question answering is one such task in which questions seemingly require multiple reasoning steps to answer. To accurately answer a multi-hop question, one must start by decomposing the given multi-hop question into simpler sub-questions, then try to answer them respectively, and finally aggregate together the information obtained from all the sub-questions. For instance, consider the multi-hop question ``What year did the band that sang `With Or Without You' form?''. To answer the question, one must first figure out the band that sang that song from one context paragraph and then find the year in which that band formed from another one. A typical approach to multi-hop QA systems is to automatically decompose the question into sub-questions, answer those questions, and then synthesize the answers to the sub-questions to answer the original question~\cite{min-etal-2019-multi,perez-etal-2020-unsupervised,khot-etal-2021-text}.

From the perspective of explainable NLP, we explore the utility of multi-hop decompositions to create explanations. One role of explanations is to help users construct a {\em mental model} of the underlying system~\cite{Chandrasekaran2017ItTT,Chakraborti2019PlanEA,Jacovi2022DiagnosingAE}. In doing so, users will be better equipped to know when the system answers can be trusted. This is especially important for large, general-purpose QA systems that can answer a wide range of questions but might have greater competencies when answering questions about some topics versus others. We hypothesize that question decompositions used to {\em probe} a neural QA model can improve users' abilities to predict whether the QA system will answer the original question correctly or not. 

Khot~\shortcite{khot-etal-2021-text} observed that improved decompositional reasoning chains for multi-hop QA correlate with increased user perceptions of trust, understandability, and preference. While perceptions of trust are important, it is also important that the trust is {\em appropriately calibrated}~\cite{MUIR1987527, DZINDOLET2003697, leetrust2004, zhang2020trust, Perkins2021TrustCA}. That is, the user should trust the system when it is worthy of that trust. For general-purpose question-answering systems built upon large-scale language models, the ability to accurately answer a question is likely to be variable based on the specific question asked.

How does a user know when to trust a QA system's answer to a particular question? If just presented with an answer, one has no cues from which to make an assessment. End-to-end QA systems that generate answers and explanations are trained to justify the answer as opposed to provide evidence of the system's competencies on a topic.

We introduce {\em probing} as an explanation strategy that helps a user determine whether to trust an answer. Probing is a process whereby a model is provided similar inputs to determine if its performance is stable when handling related inputs. In this work, we show that exposing the decomposition probes and answers to the probes to users can increase their ability to predict whether the system will answer the original question correctly. This indicates that users---without knowing the particulars of the underlying QA system---are receiving {\em actionable} cues from which to model the behavior of the system. Instead of asking for subjective perceptions of the overall system, we objectively measure the effect of the probes on instance-level interactions.

To the best of our knowledge, this paper is the first to show that probing can have a measurable impact on users in multi-hop QA. These results are also complementary to Tang et al.~\shortcite{tang-etal-2021-multi} who use decompositions to assess whether multi-hop QA systems successfully go through multiple hops when answering questions. In summary, our main findings are:
\begin{enumerate}[noitemsep]
    \item Decomposition is an effective form for probing neural QA models.
    \item Explanation created by probing the neural QA model with question decompositions can help human construct a mental model on which they can rely to predict the model behavior.
    \item Quality of decompositions matters---from the explainability perspective, existing question decomposers still have a long way to go.
\end{enumerate}

\noindent
A summary of our method is given in \autoref{fig:summary}.

\section{Preliminaries}

\subsection{Dataset}\label{ssec:datasets}

We use a popular English question-answering / reading comprehension task designed to test multi-hop reasoning: \textsc{HotPotQA} \cite{yang-etal-2018-hotpotqa}. Examples are given in \autoref{tab:hotpotqa-example}. The \textsc{HotPotQA} task involves answering questions by finding information over multiple Wikipedia articles.\footnote{We make simplifying assumptions for this task, detailed in \sect{ssec:models}.}

\begin{table}[t!]
\centering
\small
\begin{tabular}{p{0.90\linewidth}}
\toprule
\textbf{Context:} Learning, Inc. is an educational software and hardware company co-founded in 1999 by Texas businessman Neil Bush and a year later Ken Leonard. He is the fourth of six children of former President George H. W. Bush and Barbara Bush (née Pierce). \\
\textbf{Question:} Who is the mother of the Texas business man that co-founded Ignite! Learning, Inc?\\
\textbf{Answer:} Barbara Bush\\ \\
\textbf{Sub-question 1:} Who is the Texas business man who co founded Ignite Learning, Inc?\\
\textbf{Answer:} Neil Bush\\
\textbf{Sub-question 2:} Who is Neil Bush's mother?\\
\textbf{Answer:} Barbara Bush\\

\bottomrule
\end{tabular}
\caption{Example from the validation set of \textsc{HotpotQA} \cite{yang-etal-2018-hotpotqa}, as well as the associated silver question decompositions from \citet{khot-etal-2021-text}.
}
\label{tab:hotpotqa-example}
\end{table}

\subsection{Question Decompositions}\label{ssec:decomp_datasets}

As a source of high-quality question decompositions and answers, we use the sub-questions and answers provided for a subset of the \textsc{HotpotQA} validation set by \citet{khot-etal-2021-text}.\footnote{\url{https://github.com/allenai/modularqa}} These sub-questions are generated using distant supervision in the form of task-specific hints to a \textsc{BART-Large} \cite{lewis-etal-2020-bart} model trained to generate questions in the \textsc{SQuAD 2.0} dataset \cite{rajpurkar-etal-2018-know}. The answers are generated by a \textsc{RoBERTa-Large} \cite{liu2019roberta} model trained on \textsc{SQuAD 2.0}. These silver sub-question-and-answer pairs are relatively high-quality, in that the authors are able to use them to train a next-question generator that achieves high task performance on \textsc{HotPotQA} as part of a larger modular system.

All instances in the validation set have at least two sub-questions; certain questions have a third math operation sub-question that we abandon as this format only suits models with a numerical reasoning module (i.e., not conducive to being asked as a probe to the language model). The authors sampled 5 chains of sub-questions for each instance and filtered out noisy ones; we select the first from the remaining chains for our probes, and find that overall, the sub-questions and answers are of high quality and do not vary much across samples. This results in 676 instances for \textsc{HotPotQA} that have a silver question decomposition. Examples of question decompositions are given in \autoref{tab:hotpotqa-example}.

Our choice of tasks is motivated by two factors: the existence of high-quality question decompositions and answers, and the task labels are not limited to predefined categories (such as \emph{yes/no}), which limits the outputs that a fine-tuned generative model can produce when probed with sub-questions (i.e., if the dataset only contains yes/no questions, a model trained on it is unlikely to be able to answer sub-questions with anything other than yes/no). Future work can focus on extending fine-tuning protocols to apply the sub-question probing method to datasets with categorical labels.

\subsection{Models}\label{ssec:models}

We fine-tune two popular pretrained models to perform the multi-hop QA tasks: \textsc{T5-Base} \cite[220M parameters;][]{raffel2020exploring} and \textsc{BART-Base} \cite[140M parameters;][]{lewis-etal-2020-bart}. Both models are built on text-to-text encoder-decoder Transformer \cite{vaswani2017attention} architectures pretrained with denoising objectives. Both models treat question-answering tasks as generation tasks, making them well-suited for probing since they can thus also answer sub-questions in free-form natural language (rather than predicting from a fixed set of classification labels). We fine-tune the models using standard cross-entropy loss to generate the answer given the question and context. While one subtask for \textsc{HotPotQA} is to \emph{select} the relevant context, i.e., the supporting paragraph from which to extract an answer, we focus on general architectures that are not designed for retrieval. Therefore, we provide the gold context paragraph as input. More details, including input-output formatting, are given in \autoref{appendix:train}.

The \textsc{HotPotQA} leaderboard relies on two metrics for determining answer correctness, originally from SQUAD \cite{rajpurkar-etal-2016-squad}: exact match (EM), whether a prediction and the ground-truth answer matches exactly, and F1 score, the (macro-averaged) token-level overlap between a prediction and the ground-truth answer (treating both as a bag-of-tokens). Using gold context paragraphs, our models achieve comparable performance to standard baselines on the answering task, reported in \autoref{table:main_task_results}. \textsc{T5} outperforms \textsc{BART} on both metrics. Our goal is {\em not} to build the best model but to establish a model with sufficient performance on questions and sub-questions to test our hypotheses about the effect of question decompositions as explanations.

\begin{table}
\centering
\small
\begin{tabular}{llrrrrr}
    \toprule
    & \multicolumn{2}{c}{\textbf{Metric}} & \multicolumn{3}{c}{\textbf{Metric (On Subset)}}\\
    \cmidrule(lr){2-3} \cmidrule(lr){4-6}
    \textbf{Model} & \textbf{EM} & \textbf{F1} & \textbf{EM} & \textbf{Manual} & \textbf{F1}\\
    \midrule
    \textsc{T5} & $66.73$ & $79.97$ & $70.27$ & $91.27$ & $85.41$\\
    \textsc{BART} & $62.21$ & $76.18$ & $65.98$ & $88.31$ & $82.12$\\
    \bottomrule
\end{tabular}
\caption{Task performance of pretrained models on the validation set and a subset of it (see \sect{ssec:subq-probing}). ``Manual'' indicates our manual annotation for answer correctness, which is more accurate than EM. A comparable model on \textsc{HotPotQA} \cite{tu2020select} achieves $61.32$ EM and $74.81$ F1 on the full validation set.
}
\label{table:main_task_results}
\end{table}

Crucially, we never fine-tune on sub-questions. This allows our probing method to represent what a model that is \emph{only trained for the task} knows about the task, without introducing any new information that may shift the predictions of the model in favor of the explanation fine-tuning corpus \cite[such as is done in prior work;][]{roberts-etal-2020-much}.

\subsection{Probing with Sub-Questions}
\label{ssec:subq-probing}

Given the fine-tuned models, we probe on a subset of the validation instances for which we have silver sub-questions---676 instances for \textsc{HotPotQA}. This is done at inference-time following the same format as the main task, i.e., by feeding each sub-question for an instance with the instance's gold context as input to the trained model. For each instance in the dataset, this process results in a tuple of the form: main question (\textsc{Q}), gold context paragraph (\textsc{C}), the model's predicted answer to the main question (\textsc{A}), two silver sub-questions (\textsc{Sub-Q$_1$} and \textsc{Sub-Q$_2$}), and the model's predicted answers to the sub-questions (\textsc{Sub-A$_1$} and \textsc{Sub-A$_2$}).

To avoid bias introduced by requiring an exact token match or determining an F1 cutoff for correctness of a model answer, we manually annotate the instances (both main and sub-questions) for correctness. This leads to a slight increase in accuracy due to instances where EM=0 but we determine the predicted answer to be correct (e.g., the correct answer is ``Nashville'', and the model predicted ``Nashville, Tennessee''). For an example of how accuracy numbers change as the result of manual annotation, see the 4\textsuperscript{th} and 5\textsuperscript{th} columns of \autoref{table:main_task_results}.

\subsection{Simulatability}\label{ssec:human_exp_method}

To understand how \emph{faithful} explanations are to the underlying model as reflected by the mental model humans can develop of a machine learning system, the explainable AI community has long turned to \textbf{simulatability} experiments \cite[][\emph{inter alia}]{kim2016examples, doshi2017towards, ribeiro2018anchors, nguyen-2018-comparing, chandrasekaran-etal-2018-explanations, hase-bansal-2020-evaluating}. \citet{doshi2017towards} define ``forward simulation/prediction'' as the task by which ``humans are presented with an explanation and an input, and must correctly simulate the model’s output''. They class this as a form of \emph{human-grounded evaluation}, which has strengths over automatic evaluation methods because it investigates the understanding of real human users, and thus tests the utility of explanations in settings closer to true applications. Simulatability, to date, is one of the only human-grounded evaluation methods that tests the \emph{interpretability} of explanation methods rather than \emph{human preferences}, and is the most widely used due to its versatility.

We design a simulatability experiment to judge the quality of explanations. Here, we define quality as fidelity to the underlying model \cite{wiegreffe-pinter-2019-attention, jacovi-goldberg-2020-towards} and information content that provides sufficient insight into the underlying model. 

Our studies are performed using the Prolific crowdsourcing platform.\footnote{\url{https://www.prolific.co/}} These studies were approved by our institution's Institutional Review Board (IRB). We randomly select a subset of dataset instances from the 676 \textsc{HotPotQA} validation instances with silver decompositions. Participants are paid at \$15/hour, and we qualify participants by first giving them a qualification question and verifying answers manually. We require participants to be located in the U.S. and to speak English as a first language. For each set of experiments, we source a distinct set of participants (no overlap) to avoid any bias in annotations that could occur from seeing past versions of the task or questions. For all experiments, we report Fleiss' $\kappa$ \cite{fleiss1971measuring} for binary or nominal data, and Kendall's $\tau$ \cite{kendall1938new} for ordinal data.

For performance metrics, we report accuracy, F1, precision, recall, and Matthew's Correlation Coefficient (MCC; also known as the {\em phi} coefficient outside machine learning).

\section{Sub-question answering can distinguish incorrect and correct model predictions}

We first investigate the extent to which performance on sub-question-answering is tied to performance on the main QA tasks. We split the validation set instances into two groups: those for which the model predicts the answer for the main question correctly, and those for which it does not. Results are presented in \autoref{table:split_results}, which suggests sub-question accuracy is indicative of model performance, with a meaningful difference in sub-question accuracy observed between the instances which the model predicts correctly vs. those it does not.

\begin{table}
\centering
\small
\begin{tabular}{lllr}
    \toprule
    & \textbf{Model} & & \textbf{Sub-Q}\\
    \textbf{Model} & \textbf{Pred.} & \textbf{n} & \textbf{Accuracy}\\
    \midrule
    \textsc{T5} & Correct & $617$ & $\mathbf{85.09}$\\
    & Incorrect & $59$ & $64.41$\\
    \textsc{BART} & Correct & $597$ & $\mathbf{85.59}$ \\
    & Incorrect & $79$ & $60.76$ \\
    \bottomrule
\end{tabular}
\caption{Combined sub-question task performance, split by whether the model predicted the main question correctly or not.}
\label{table:split_results}
\end{table}

\section{Sub-question explanations allow humans to predict model behavior}

\label{sec:simulatability}

Given the correlations between model's performance on main QA and sub-QA, we take a step forward to ask: can humans gain any useful insight from such correlations? We perform a simulatability experiment to measure how well the sub-question explanations can help humans predict model behaviors on the main \textsc{HotPotQA} task.

To this end, we design and conduct a human participant study to investigate crowd annotators' ability to make accurate predictions about model performance given question decompositions as explanations, following the protocol given in \sect{ssec:human_exp_method}. We select a random 100-instance sample from the 676 \textsc{HotPotQA} validation instances, balanced such that the model predicts 50 instances correctly and 50 incorrectly, and perform the probing procedure described in \sect{ssec:subq-probing} on the best-performing model (\textsc{T5-Base}), which results in tuples of the form (\textsc{Q}, \textsc{C}, \textsc{A}, \textsc{Sub-Q$_1$}, \textsc{Sub-A$_1$}, \textsc{Sub-Q$_2$}, \textsc{Sub-A$_2$}), where all answers are predicted by the model.

Our goal is to observe how much the \textsc{Sub-QA} explanations help human annotators predict model behavior over a baseline that does not include these explanations, as well as investigate how the context (\textsc{C}) \& the predicted answer (\textsc{A}) could potentially impact human's performance of diagnosing model errors. We design five different settings in which human participants are provided with different combinations of information. After reading the combination of information we present, the participants are asked to make their predictions about model's behavior on the main question (\textsc{Q}), i.e., whether or not the model will be able to correctly answer the given \textsc{Q}.

We recruit 50 participants on Prolific and split them into 5 batches, each of which contains 10 participants. Given the 100-instance sample, we split it into 5 batches of 20 questions each. We follow a Latin Square design, similarly to \cite{gonzalez2020reverse}, to ensure that each group of participants only sees each set of questions under one condition: (\textsc{Q}, \textsc{A}), (\textsc{Q}, \textsc{A}, \textsc{Sub-Q}), (\textsc{Q}, \textsc{A}, \textsc{Sub-Q}, \textsc{Sub-A}), (\textsc{Q}, \textsc{Sub-Q}, \textsc{Sub-A}), or (\textsc{Q}, \textsc{C}, \textsc{Sub-Q}, \textsc{Sub-A}), yet each condition is tested on both all 50 annotators and all 100 questions. This ensures that no bias in human predictions occurs due to having previously seen the questions and model predictions. Example of the UI that participants see is given in \autoref{fig:human_study_1+2}. Finally, we collect their predictions and compute the performance scores using the actual main question's answer correctness as the ground truth.

\begin{table*}
\centering
\small
\begin{tabular}{lrrrrr}
    \toprule
    & \multicolumn{5}{c}{\textbf{Metric}}\\
    \cmidrule(lr){2-6}
    \textbf{Setting} & \textbf{Acc.} & \textbf{F1} & \textbf{Precision} & \textbf{Recall} & \textbf{MCC}\\
    \midrule
    \textsc{(Q, A)} & $58.17_{1.55}$ & $65.74_{1.63}$ & $55.36_{1.59}$ & $\textbf{83.48}_{2.29}$ & $19.15_{3.40}$\\
    \textsc{(Q, A, Sub-Q)} & $56.57_{1.20}$ & $62.94_{1.82}$ & $53.78_{1.71}$ & $79.32_{2.80}$ & $15.41_{2.87}$\\
    \textsc{(Q, A, Sub-Q, Sub-A)} & $\textbf{63.50}_{1.39}^*$ & $\textbf{68.95}_{1.15}$ & $\textbf{60.82}_{1.46}$ & $82.12_{1.60}$ & $\textbf{29.50}_{2.92}$\\
    \hline
    \textsc{(Q, Sub-Q, Sub-A)} & $53.07_{1.43}$ & $61.25_{1.54}$ & $52.49_{1.71}$ & $76.88_{2.26}$ & $8.29_{3.13}$\\
    \textsc{(Q, C, Sub-Q, Sub-A)} & $57.00_{1.66}$ & $64.61_{1.62}$ & $54.82_{1.72}$ & $80.37_{1.78}$ & $14.79_{3.67}$\\
    \bottomrule
\end{tabular}
\caption{Simulatability performance of human participants on 100 validation instances of \textsc{HotPotQA} given different input combinations. The majority baseline for accuracy is 50.00 since the dataset is fully balanced. All the statistics are computed by averaging across 50 participants, with standard errors included in subscripts. $*$: The setting's accuracy score distribution over 50 annotators is statistically significantly different from \emph{all other methods} at $p=0.01$ using two-sided Mann-Whitney $U$ tests.}
\label{table:human_studies_result_1}
\end{table*}

Results are presented in \autoref{table:human_studies_result_1}. The average inter-annotator agreement is $\kappa = 0.24$. In order to ensure that human users are not simply performing the \textsc{HotPotQA} labelling task themselves, we validate this by first providing users with (\textsc{Q}, \textsc{A}) pairs, asking them ``Do you think the answer to the given multi-hop question provided by the question-answering system is correct?''. Because they are not given the context, \textsc{C}, this serves as a lower bound in quantifying any biases the participants may have about AI systems.

We apply the two-sided Mann-Whitney $U$ test \cite{mann1947test} for statistical significance on accuracy numbers. Participant accuracy given (\textsc{Q}, \textsc{A}, \textsc{Sub-Q}, \textsc{Sub-A}) is statistically significantly different at $p=0.01$ from all other settings, and results in substantially higher performance across all metrics except recall. This demonstrates that our proposed \textsc{Sub-QA} explanation method does help humans make more accurate predictions about model behavior on the main question (\textsc{Q}) than simply seeing model predictions (\textsc{Q}, \textsc{A}). We additionally validate that both sub-questions and sub-answers are important---when we ablate sub-answers, humans do poorly at the simulatability task given (\textsc{Q}, \textsc{A}, \textsc{Sub-Q}), resulting in no significant performance difference over (\textsc{Q}, \textsc{A}) pairs.

Having the answer (\textsc{A}) greatly improves the prediction performance, whereas the context (\textsc{C}) does not significantly impact human's prediction performance. Meanwhile, the proved feasibility of human's making accurate prediction about model behavior using \textsc{Sub-QA} explanations suggests a potential future direction for establishing an alternative for carrying out real annotation activities in order to diagnose QA system's error. The benefit of such alternative is obvious: humans will no longer have to conduct the question decomposition and perform the actual multi-hop reading comprehension by themselves. Instead, they may solely rely on or at least gain useful insights from their mental model about the QA system to save time and effort when trying to diagnose the error.

\section{Quality of question decompositions matters}

\label{sec:decompositions}

Prior work has shown that predictions from question decomposition models can improve task performance on \textsc{HotPotQA} when being part of a larger modular system \cite{min-etal-2019-multi, perez-etal-2020-unsupervised, khot-etal-2021-text}, but qualitative inspection reveals a lack of quality in many cases. To investigate whether such sub-question-generation models can provide interpretability, we explore the effect of sub-question quality on utility of question decompositions as explanations in our probing setup. Namely, we conduct simulatability experiments and measure performance variation in humans' ability to guess model predictions based on the quality of the \textsc{Sub-QA} explanations they received.

We use decomposition predictions from three trained question decomposers developed as part of larger modular QA systems in prior work: 
a) \textsc{ModularQA} \cite{khot-etal-2021-text}; b) One-to-N Unsupervised Sequence transduction \cite[\textsc{ONUS}; ][]{perez-etal-2020-unsupervised}; and c) \textsc{DecompRC} \cite{min-etal-2019-multi}. \textsc{ModularQA} is a  next-question-prediction \textsc{BART-Large} model trained on the silver decompositions described in \sect{ssec:decomp_datasets}. \textsc{ONUS} is trained to decompose complex questions from the internet into simpler questions using supervision from noisy pseudo-decompositions. \textsc{DecompRC} is trained on a mix of supervision and heuristics to create sub-questions from the tokens in the original question, framing the task as span prediction. Examples of the question decompositions produced by each method are in \autoref{tab:question_decomposition_examples_for_4_decomposers}. \textsc{ONUS} and \textsc{DecompRC} always produce two sub-questions; \textsc{ModularQA} follows the form of \textsc{Silver} and thus also results in 2 sub-questions per-instance once math operations are removed (\sect{ssec:decomp_datasets}).

\begin{table*}[t!]
\centering
\small
\begin{tabular}{p{0.90\linewidth}}
\toprule
\textbf{\textsc{Silver} \cite{khot-etal-2021-text}} \\
\midrule[0.03em]
\textbf{Sub-question 1:} During what war was Pavillon du Butard occupied by the Prussians?\\
\textbf{Sub-question 2:} What was the name that the French called the Franco-Prussian War?\\

\toprule[0.03em]
\textbf{\textsc{ModularQA} \cite{khot-etal-2021-text}} \\
\midrule[0.03em]
\textbf{Sub-question 1:} During what war was the Pavillon du Butard occupied?\\
\textbf{Sub-question 2:} What is the French name for the Franco-Prussian War?\\

\toprule[0.03em]
\textbf{\textsc{ONUS} \cite{perez-etal-2020-unsupervised}} \\
\midrule[0.03em]
\textbf{Sub-question 1:} What is the name the french give to the war?\\
\textbf{Sub-question 2:} During which war did the prussians occupy the pavillon du butard?\\

\toprule[0.03em]
\textbf{\textsc{DecompRC} \cite{min-etal-2019-multi}} \\
\midrule[0.03em]
\textbf{Sub-question 1:} which war during which the prussians occupied the pavillon\\
\textbf{Sub-question 2:} what is the name the french give to Franco-Prussian War du butard?\\

\bottomrule
\end{tabular}
\caption{Examples of the question decompositions produced by \{\textsc{Silver}, \textsc{ModularQA}, \textsc{ONUS}, \textsc{DecompRC}\} for the question ``What is the name the French give to the war during which the Prussians occupied the Pavillon du Butard?''.
}
\label{tab:question_decomposition_examples_for_4_decomposers}
\end{table*}

We repeat the crowdsourcing process in \sect{sec:simulatability}, randomly sampling a subset of 30 correctly-predicted instances and 30 incorrectly-predicted instances from the 100 selected in \sect{sec:simulatability}. We probe the \textsc{T5} model with \textsc{Sub-Q$_1$} and \textsc{Sub-Q$_2$} produced by each of the 4 sources: \{\textsc{Silver}, \textsc{ModularQA}, \textsc{ONUS}, \textsc{DecompRC}\}, and collect its \textsc{Sub-A$_1$}, \textsc{Sub-A$_2$} responses. Tuples of (\textsc{Q}, \textsc{A}, \textsc{Sub-Q$_1$}, \textsc{Sub-A$_1$}, \textsc{Sub-Q$_2$}, \textsc{Sub-A$_2$}) are presented to 30 new annotators (who have not participated in previous experiments) following the setup in \sect{sec:simulatability}.

Similar to \sect{sec:simulatability}, we perform a Latin Square design by equally splitting the participants and the questions into 3 batches, such that each participant group only observes each subset of questions under one experimental condition (either \textsc{ModularQA}, \textsc{ONUS}, or \textsc{DecompRC} predictions). Annotator performance metrics at predicting answer correctness, averaged across all 30 participants, are presented in \autoref{table:human_studies_result_2}, along with annotator performance given \textsc{Silver} sub-questions. The average inter-annotator agreement is $\kappa = 0.29$.

\begin{table*}
\centering
\small
\begin{tabular}{lrrrrr}
    \toprule
    & \multicolumn{5}{c}{\textbf{Metric}}\\
    \cmidrule(lr){2-6}
    \textbf{Decomposer} & \textbf{Acc.} & \textbf{F1} & \textbf{Precision} & \textbf{Recall} & \textbf{MCC}\\
    \midrule
    \textsc{Silver} & $\textbf{63.50}_{1.39}^*$ & $\textbf{68.95}_{1.15}$ & $\textbf{60.82}_{1.46}$ & $\textbf{82.12}_{1.60}$ & $\textbf{29.50}_{2.92}$\\
    \hline
    \textsc{ModularQA} & $58.33_{1.94}$ & $63.19_{2.06}$ & $59.11_{1.54}$ & $69.22_{3.07}$ & $16.23_{4.22}$\\
    \textsc{DecompRC} & $57.67_{1.51}^*$ & $60.90_{1.79}$ & $60.33_{1.55}$ & $64.61_{3.34}$ & $15.64_{3.07}$\\
    \textsc{ONUS} & $53.11_{1.53}$ & $52.80_{2.70}$ & $55.96_{1.53}$ & $54.13_{4.05}$ & $6.07_{3.22}$\\
    \bottomrule
\end{tabular}
\caption{Simulatability performance of human participants on 60 validation instances of \textsc{HotPotQA}, where \textsc{Sub-Q} are provided by different question decomposers and \textsc{Sub-A} are obtained from our \textsc{T5-Base} model. The majority baseline for accuracy is 50.00 since the dataset is fully balanced. All but \textsc{Silver} statistics (copied from \autoref{table:human_studies_result_1}) are computed by averaging across 30 participants, with standard errors included in subscripts. $*$: The method's accuracy score distribution over 30 annotators is statistically significantly different from all the methods below it at $p=0.05$ using two-sided Mann-Whitney $U$ tests.
}
\label{table:human_studies_result_2}
\end{table*}

We apply the two-sided Mann-Whitney $U$ test \cite{mann1947test} for statistical significance on accuracy numbers. Human performance scores from the trained decomposers are all worse than the \textsc{Silver} decomposer at a statistically-significantly different level ($p=0.05$). This indicates that there are still notable gaps between the quality of \textsc{Silver}'s and other existing decomposers' \textsc{Sub-QA} explanations. \textsc{ONUS} question decompositions consistently provide the least explanatory power. Despite \textsc{DecompRC}'s methodological simplicity, the explanatory power of its question decompositions is comparable to \textsc{ModularQA}, though \textsc{ModularQA} is the highest-performing predictive model overall. This is further supported by statistical significance results, which reveal that both \textsc{ModularQA} and \textsc{DecompRC} are statistically-significantly different from \textsc{ONUS}, but not from one another ($p=0.05$).

To further investigate the quality differences of different sources of question decompositions as measured by \emph{human preferences}, we conduct an additional study where participants are asked to rank sources of \textsc{Sub-QA} explanations based on their quality. Specifically, we again recruit 30 new participants and each of them is asked to rank four decomposers' \textsc{Sub-QA} explanations for 30 random question samples in terms of three criteria: well-formedness, relatedness, and informativeness. Example of the UI that participants see is given in \autoref{fig:human_study_3}. Results are presented in \autoref{table:human_studies_result_3}. Inter-annotator agreement, as measured by Kendall's Tau \cite{kendall1938new}, is $\tau = 0.32$. \textsc{Silver} decomposer is consistently preferred under all measurement criteria; \textsc{ModularQA} is consistently second-best, followed by \textsc{ONUS} and \textsc{DecompRC}. This also echoes results reported in \citet{khot-etal-2021-text} (who only compared \textsc{ModularQA} to \textsc{DecompRC}).

\begin{table*}
\centering
\small
\resizebox{\textwidth}{!}{
\begin{tabular}{l|rrrr|rrrr|rrrr}
    \toprule
    \multicolumn{1}{c}{} & \multicolumn{4}{c}{\textbf{Well-formedness}} & \multicolumn{4}{c}{\textbf{Relatedness}} & \multicolumn{4}{c}{\textbf{Informativeness}}\\
    \cmidrule(lr){2-5} \cmidrule(lr){6-9} \cmidrule(lr){10-13}
    \textbf{Decomposer} & \textbf{1st} & \textbf{2nd} & \textbf{3rd} & \textbf{4th} & \textbf{1st} & \textbf{2nd} & \textbf{3rd} & \textbf{4th} & \textbf{1st} & \textbf{2nd} & \textbf{3rd} & \textbf{4th}\\
    \midrule
    \textsc{Silver} & $\textbf{50.2}_{3.3}$ & $37.3_{2.9}$ & $6.4_{1.3}$ & $6.1_{1.2}$ & $\textbf{41.8}_{3.0}$ & $\textbf{37.6}_{2.7}$ & $11.5_{1.4}$ & $9.1_{1.3}$ & $\textbf{41.7}_{3.0}$ & $37.4_{2.8}$ & $11.7_{1.3}$ & $9.2_{1.3}$\\
    \textsc{ModularQA} & $35.5_{2.8}$ & $\textbf{45.1}_{3.5}$ & $12.0_{1.9}$ & $7.4_{1.5}$ & $37.0_{2.5}$ & $36.7_{2.9}$ & $14.9_{1.6}$ & $11.4_{1.3}$ & $36.5_{2.5}$ & $\textbf{38.3}_{2.9}$ & $14.2_{1.8}$ & $11.0_{1.4}$\\
    \textsc{ONUS} & $9.1_{1.3}$ & $13.2_{2.0}$ & $\textbf{54.9}_{4.5}$ & $22.8_{2.2}$ & $14.1_{1.3}$ & $18.6_{1.6}$ & $\textbf{37.8}_{3.6}$ & $29.5_{2.2}$ & $15.7_{1.5}$ & $17.0_{1.6}$ & $\textbf{40.1}_{3.4}$ & $27.2_{2.0}$\\
    \textsc{DecompRC} & $5.2_{1.5}$ & $4.4_{1.0}$ & $26.7_{2.7}$ & $\textbf{63.7}_{3.8}$ & $7.1_{1.3}$ & $7.1_{1.2}$ & $35.8_{2.4}$ & $\textbf{50.0}_{3.2}$ & $6.1_{1.3}$ & $7.3_{1.2}$ & $34.0_{2.5}$ & $\textbf{52.6}_{3.1}$\\
    \bottomrule
\end{tabular}
}
\caption{Percentages (\%) of the time each decomposer is listed in a ranking spot. Human participants rank all four question decomposers in terms of the well-formedness, relatedness, and informativeness of their corresponding questions and answers. Each annotator judges the same 30 instances, and results are averaged across 30 annotators. Subscripts indicate standard errors over 30 annotators.}
\label{table:human_studies_result_3}
\end{table*}

\section{Related Work}
\label{sec:related_work}

Multiple prior works have concluded that question-answering as a \emph{form} \cite{gardner2019question} is a good choice for probing pretrained models \cite{roberts-etal-2020-much, marasovic2021few}. \citet{roberts-etal-2020-much} fine-tune a model on a dataset of questions and answers, but claim this does not introduce new information to the model and only teaches form for effective QA probing. However, this claim is not well-supported, as fine-tuning removes any guarantees that the questions answered at test-time reflect information learned during pre-training alone, and is not zero- or few-shot. We avoid this by performing probing in a truly zero-shot manner (i.e., \textbf{we never fine-tune on sub-questions}). Additionally, the method of \citet{roberts-etal-2020-much} does not probe for instance-level prediction explanations; the authors instead use a fixed set of questions on general topics. In our work, we use the instance-level explanations we obtain from probing with sub-questions to test whether these explanations give humans an accurate mental model of the system \cite{Jacovi2022DiagnosingAE}.

Most related to our work is that of \citet{tang-etal-2021-multi}, who also investigate whether model architectures for multi-hop QA can answer single-hop questions. They find that there is a significant percentage of questions for which the model answers the main question correctly, but cannot correctly answer the corresponding single-hop sub-questions. However, because they use a model to produce question decompositions, their results may be confounded by errors or low quality of the questions themselves, which our work circumvents by using a silver source of sub-questions; we also investigate the effect of sub-question quality on the final results.

\section{Conclusions}
\label{sec:conclusion}

We have demonstrated the utility of question decomposition as an effective means to probe pre-trained multi-hop question-answering models for supporting evidence. Through simulatability experiments, we show the effectiveness of this explanation form at allowing humans to predict model behavior, a sign that it helps humans to form an accurate \emph{mental model} of the machine learning system \cite{Jacovi2022DiagnosingAE}. This ability to predict system performance occurs at the instance level instead of a sense of trust of the overall system, which can be important if the accuracy of the system is variable based on the question.

Our results indicate that explanations based on decompositional probes can be beneficial to users when the sub-questions are of reasonable quality. Our analyses indicate that existing decomposition systems, however, have considerable room for improvement. We can now look at the state of research in decomposition systems not only as to whether they improve multi-hop question answering, but whether they provide users with more calibrated trust.

\section{Limitations}
\label{sec:limitations}

Our simulability study results (Section~\ref{sec:simulatability}) are conducted on silver labels. As Section~\ref{sec:decompositions} reveals, there is a need for higher-quality question decompositions. While we have demonstrated the potential for decomposition probes to help users build mental models of system behavior, these results are not fully realizable in real applications until decomposition systems improve.

The probing strategy explored in this paper is particular to the QA setting and datasets that don't have predefined categories of answers. Other probing strategies may exist that are not explored in this paper.

It is noted that multi-hop questions do not always require multi-hop reasoning to solve. Indeed we intentionally use a non-multi-hop question-answering model to answer the original question to disadvantage the system so that explanations are required. Multi-hop questions afford the use of a decompositional probing strategy. Our study did not look at non-multi-hop questions, which may require other probing strategies yet to be invented.

\section{Ethics \& Broader Impacts}
\label{sec:ethics}

All datasets used in this work are public. We did not collect any personal information from our human participants nor did we present them with any harmful model outputs.

QA systems, as with all language model based systems, are prone to unwanted biases; this is beyond the scope of our paper. QA systems present safety issues when humans act upon answers that are wrong. Our paper is a step toward helping human users understand when they should or should not trust the answers.

\section{Acknowledgements}
This work was done while SW was at the Georgia Institute of Technology. We thank members of the Entertainment Intelligence and Human-Centered AI lab at Georgia Tech for valuable feedback and discussions.

\bibliography{anthology, custom}
\bibliographystyle{acl_natbib}

\appendix

\section{Additional Details}
\label{appendix:train}

We use Huggingface Datasets \cite{lhoest-etal-2021-datasets} and Huggingface Transformers \cite{wolf-etal-2020-transformers}. Models are trained with a learning rate linearly decaying from $5E-5$, a batch size of $64$, and default values for Adam \cite{kingma2014adam}, gradient clipping, and dropout. We train for a maximum $200$ epochs, performing early stopping on the validation loss with a patience of $10$ epochs. All models are trained on an NVIDIA GeForce GTX 1080 GPU ($8$ GB memory) and on average take approximately $14$ hours to train, converging in around $12$ epochs. Input-output formatting is:

{\footnotesize
\begin{verbatim}
input_string = (f"question: {question} 
    context: {passage}")
output_string = (f"{answer}")
\end{verbatim}
}

The \textsc{HotPotQA} dataset has 90,447 train and 7,405 validation instances. In the \textsc{HotPotQA} leaderboard, there are two evaluation settings: distractor and full-wiki. In the distractor setting, models are given 10 paragraphs where 2 of them are gold paragraphs needed to answer the question and the other 8 are ``distractors''. In the full-wiki setting, models are given the first paragraphs of all Wikipedia articles without the gold paragraphs specified. We do not submit to the leaderboard and thus cannot report test set performance, since we simplify the task and pass the 2 gold context paragraphs as input directly (\sect{ssec:models}), which does not align with either evaluation setting.

\begin{figure*}
    \centering
    \includegraphics[width=\textwidth]{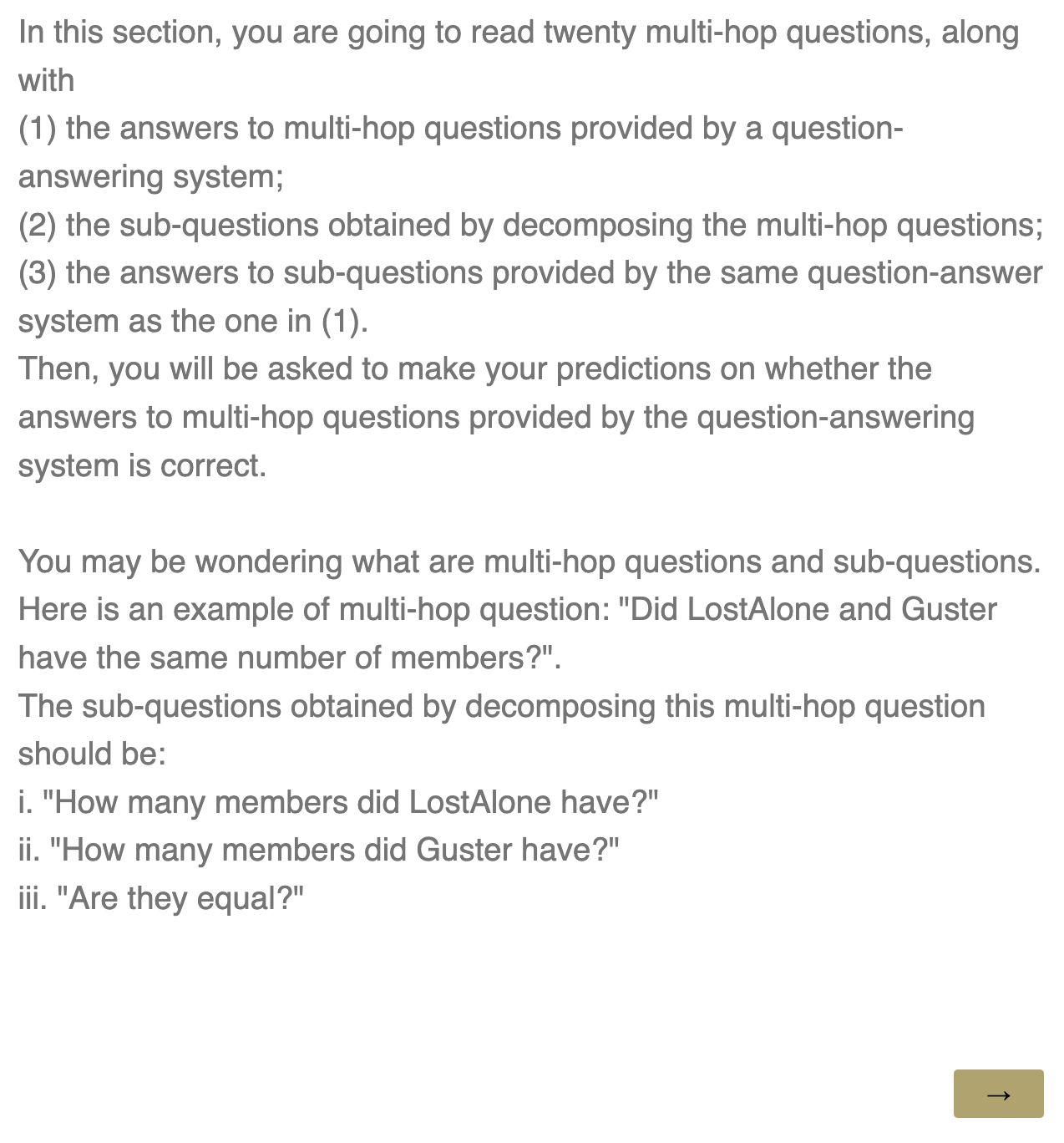}
    \caption{The Prolific interface for simulatability experiments in \autoref{sec:simulatability}.}
    \label{fig:human_study_1+2_instruction}
\end{figure*}

\begin{figure*}
    \centering
    \includegraphics[width=\textwidth]{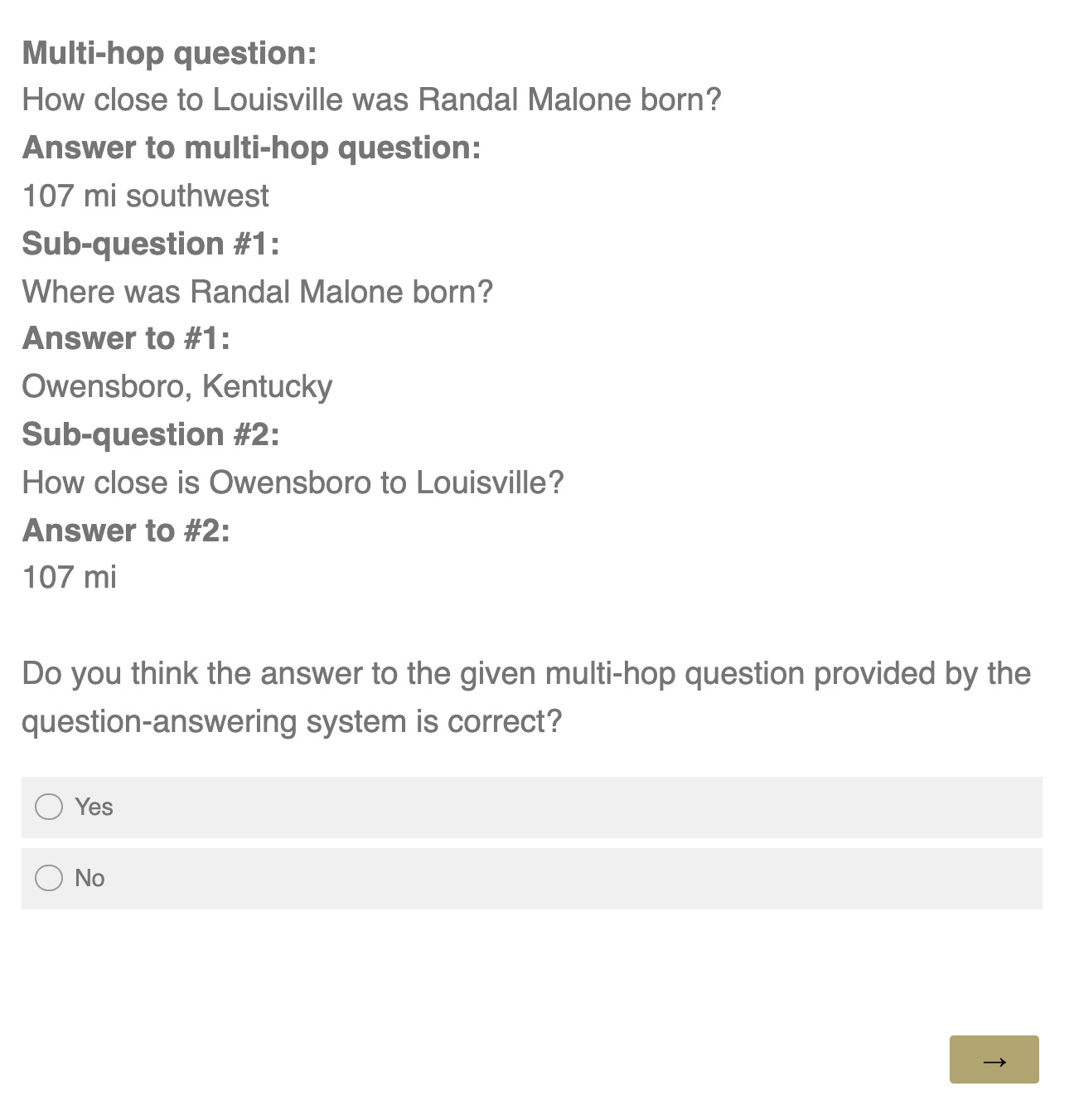}
    \caption{The Prolific interface for simulatability experiments in \autoref{sec:simulatability}.}
    \label{fig:human_study_1+2}
\end{figure*}

\begin{figure*}
    \centering
    \includegraphics[height=0.9\textheight]{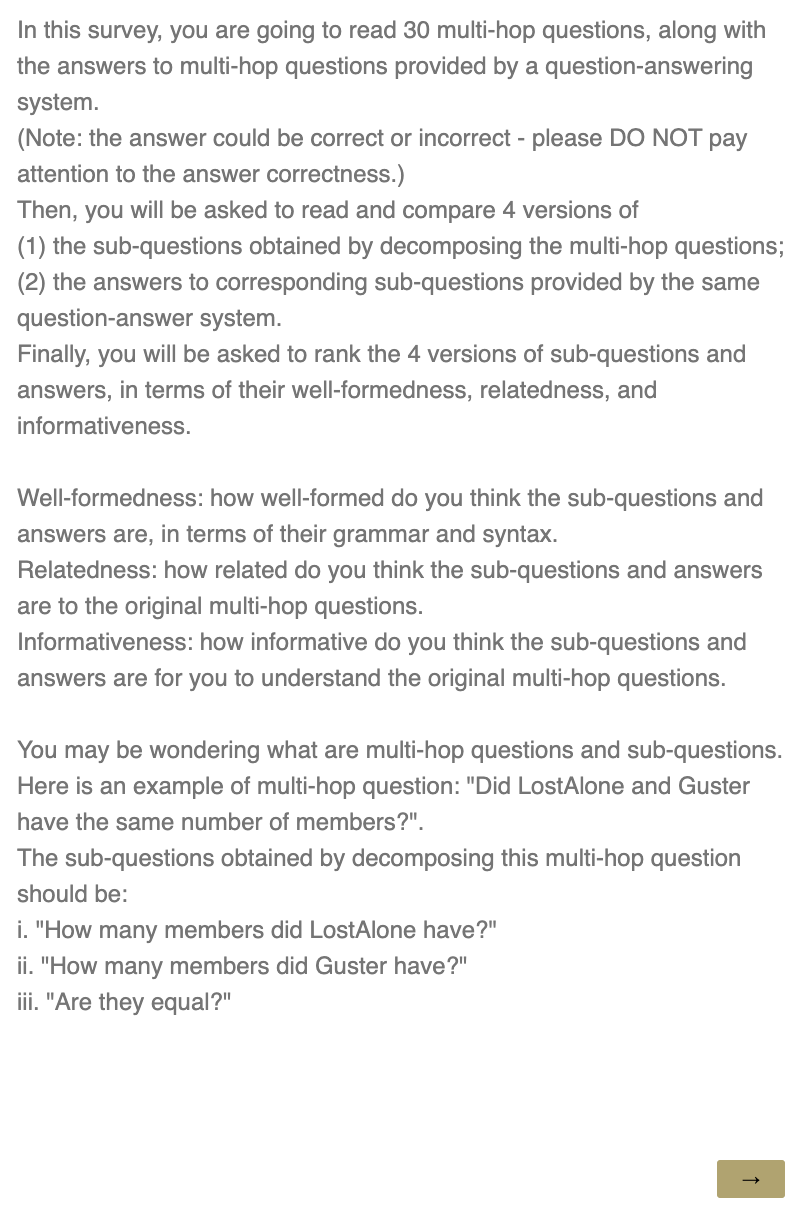}
    \caption{The Prolific interface for ranking experiments in \autoref{sec:decompositions}.}
    \label{fig:human_study_3_instruction}
\end{figure*}

\begin{figure*}
    \centering
    \includegraphics[height=0.9\textheight]{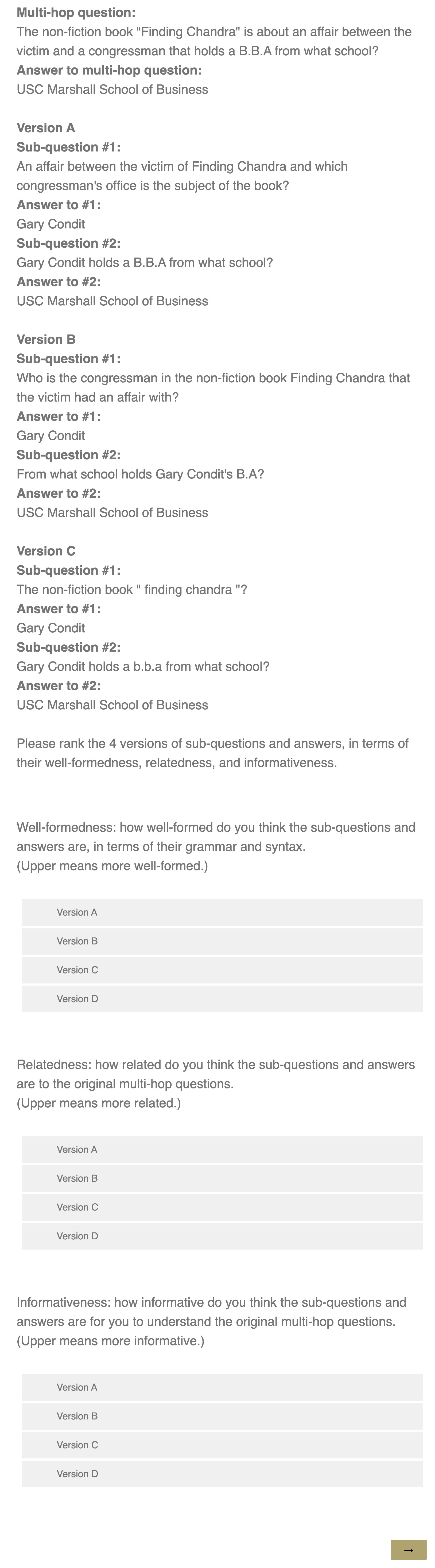}
    \caption{The Prolific interface for ranking experiments in \autoref{sec:decompositions}.}
    \label{fig:human_study_3}
\end{figure*}

The Prolific interfaces for the human participant studies conducted in \autoref{sec:simulatability} are shown in \autoref{fig:human_study_1+2_instruction} and \autoref{fig:human_study_1+2}; the Prolific interfaces for the human participant studies conducted in \autoref{sec:decompositions} are shown in \autoref{fig:human_study_3_instruction} and \autoref{fig:human_study_3}.

\end{document}